\documentclass[11pt,twocolumn,a4paper]{article}
\usepackage[hyperref]{acl2020}
\usepackage{times}
\usepackage{latexsym}
\usepackage{booktabs, graphicx, amsmath, hyperref}

\usepackage{siunitx}
\usepackage{svg}
\usepackage{microtype}

\aclfinalcopy 

\makeatletter
\let\@fnsymbol\@arabic
\makeatother

\title{ReINTEL Challenge 2020: A Comparative Study of Hybrid Deep Neural Network for Reliable Intelligence Identification on Vietnamese SNSs}

\author{Hoang Viet Trinh, Tung Tien Bui, Tam Minh Nguyen\\ \textbf{Huy Quang Dao, Quang Huu Pham, Ngoc N. Tran} \\
  \texttt{trinh.viet.hoang@sun-asterisk.com} \\
  AI Research Team, R\&D Lab, Sun* Inc. \vspace{0.5em} \\
  \textbf{Ta Minh Thanh}\\
  Le Quy Don Technical University, Ha Noi, Vietnam \\
  
}

\usepackage{indentfirst}
\begin{document}
\maketitle
\begin{abstract}
The overwhelming abundance of data has created a misinformation crisis. Unverified sensationalism that is designed to grab the readers' short attention span, when crafted with malice, has caused irreparable damage to our society's structure. As a result, determining the reliability of an article has become a crucial task. After various ablation studies, we propose a multi-input model that can effectively leverage both tabular metadata and post content for the task. Applying state-of-the-art fine-tuning techniques for the pretrained component and training strategies for our complete model, we have achieved a 0.9462 ROC-score on the VLSP private test set.
\end{abstract}

\section{Introduction}
\subsection{Overview}

The fast growth of social media and misinformed contents have posed an incremental challenge of exposing untrustworthy news to billions of their global users, including 65 million Vietnamese users \cite{digital_2020}. Consequently, the spread of mistrust information on social cites has placed real damages on government, policymakers, organizations, and citizens of many countries \cite{doi:10.1080/15205436.2020.1750656, inproceedings}, resulting in an urge for fast and large-scale fact-checking online contents. With the enormous amount of news and information on the internet daily, this is impossible to be efficiently done only by human efforts, putting a quest to create a trustworthy system to perform the task automatically.

Reliable Intelligence Identification on Vietnamese SNSs (ReINTEL) is the task of reliable or unreliable social-network-sites (SNSs) identification. The main difficulties of these tasks, including:

\begin{itemize}
    \item The given data (contents of social sites) is unstructured, containing mostly texts combined with metadata (including: images, dates, numbers, username, id, $etc$). The meta-information is partially missing and incorrect, making the usage of those data more challenging.
    \item  The problem is multi-modal learning, which `involves relating information from multiple sources' \cite{inbook}, resulting in the search for a proper combination of features from those sources to learn a unified model with high performance. 
\end{itemize}

\subsection{Our contributions}

In this paper, we propose our methods to resolve these above-mentioned problems. With thorough experiments, we determined to answers two main questions: Should we incorporate multi-source data? Furthermore, how to combine them in terms of training strategies? Our contributions are as followed:

\begin{itemize}
    \item We provide a reliable method of data cleansing, making metadata ready for prediction.
    \item More importantly, we are the first who construct a comprehensive comparative study to discover the effectiveness of models when incorporating multi-source data with different training strategies. 
    Our experiment’s results reveal that:
    
    \begin{itemize}
        \item Models using text or meta-features alone has a crucial gap in performance, indicating that texture information is significantly more predictive than metadata. 
        \item Models utilize multi-source data with different training strategies results in a wide range of performance. This finding implies that combining data in training has a significant impact on the overall performance. 
        \item Combining data from multi-sources with particular training plans leads to our best models. Additionally, the model trained with metadata alone performs significantly better than a random guess, shedding light on the meta data's informativeness. 
    \end{itemize}

    \item We apply state-of-the-art transfer learning methods for textual feature extractions and neural network (in comparison with other traditional machine learning methods) for tabular-data feature representation, achieving the competitive performance of 0.9418
    ROC-score on the public test set (ranked 2nd) and 0.9462 ROC-score (ranked 3th) on the private test set.
\end{itemize}

\subsection{Roadmap}
    
   In the following sections, we briefly review some related works involve with our methods. Next, in section 3, we illustrate our method in detail. Our experiments are described in Section 4, including dataset description, data preprocessing methods, and our model configurations, whereas Section 5 indicates all of our experimental results. Finally, section 6 is the conclusion for our proposed framework. 

\section{Related work}
\subsection{Contextual Representation For Text}

Recent works on learning universal representation for text, namely Elmo \cite{DBLP:journals/corr/abs-1802-05365}, GPT \cite{Radford2018ImprovingLU}, BERT \cite{devlin2018bert} have brought remarkable improvements for wide, diverse NLP downstream tasks: Text Classification, Question Answering and Named Entity Recognition. In contrast to traditional methods such as Word2vec \cite{mikolov2013efficient} or Glove \cite{pennington-etal-2014-glove} which learns context-independent word embeddings, universal language models were trained on a massively large amount of unlabeled data with different pretext tasks, including causal language modeling and masked language modeling, to learn a deep contextual representation of words given its context.

\subsection{Fake News Detection on SNSs}
Studies of fake news identification on social network sites have gained significant attention recently. Most of them utilize data from multiple sources. For example, CSI \cite{DBLP:journals/corr/RuchanskySL17}, a framework with several modules based on Long Short-Term Memory \cite{10.1162/neco.1997.9.8.1735} and a fully connected layer that utilizes the article's contents, the users' responses and behaviors of source users who promote it. Another instance is dEFEND \cite{10.1145/3292500.3330935}, which exploits both news contents and user comments with a deep hierarchical co-attention network to learn a rich representation for fake news detection. From a slightly different point of view, TriFN \cite{DBLP:journals/corr/abs-1712-07709} models a tri-relationship between users, publishers, and new contents by several embedding methods and experiments promising results.

Although utilizing multi-source data, existing research appears to lack a comprehensive study on the effectiveness of input-combination strategies. 

\subsection{Vietnamese Natural Language Processing}

Inspired by BERT's textual learning methods, PhoBERT \cite{nguyen2020phobert} was proposed to extend the successes of deep pre-trained language models to Vietnamese. Its pretraining approach is based on RoBERTa \cite{DBLP:journals/corr/abs-1907-11692} training strategies to optimize BERT training procedure. Additionally, PhoBERT also consists of two different settings, PhoBERT Base, which uses 12 Transformer Encoder layers and 24 layers with PhoBERT Large. It improves many Vietnamese NLP downstream tasks. For instance, Pham \cite{inproceedings} introduced novel techniques to adapt general-purpose PhoBERT to a specific text classification task and archives state of the art on Vietnamese Hate Speech Detection (HSD) campaign.

\section{Methodology}
\subsection{Dataset}

\begin{table}[]
\centering
\caption{ Statistics of the datasets. }
\begin{tabular}{lc}
\toprule & Dataset \\ 
\midrule  Total News & 5172 \\
 Users & 3706  \\
 Unique News & 5087  \\
 News have images & 1287  \\
 Reliable News & 4238  \\
 Unreliable News & 934 \\
\bottomrule
\end{tabular}
\label{table:dataset}
\end{table}

In this paper, we use the dataset provided by VLSP organizers for ReINTEL task \cite{le2020reintel}, composed of contents from Vietnamese social network sites (SNSs), e.g., Facebook, Zalo, or Lotus \cite{digital_2020}. There are approximately 5,000 labeled training examples, while the test set consists of 2,000 unlabeled examples. Each example is provided with information about the news's textual content, timestamp, number of likes, shares, comments, and attached pictures. Table ~\ref{table:dataset} indicates the detailed statistic of the dataset, the data distribution of reliable and unreliable news was heavily imbalanced and skewed toward trustworthy contents. 

\subsection{Data preprocessing}
Fake news can be studied with respect to four perspectives: (i) knowledge-based (focusing on the false knowledge in fake news); (ii) style-based (concerned with how fake news is written); (iii) propagation-based (focused on how fake news spreads); and (iii) credibility-based (investigating the credibility of its creators and spreaders) \cite{DBLP:journals/corr/abs-1812-00315}. In this task, with the ReINTEL dataset, we focused on knowledge-based and credibility-based. Specifically, we performed the following preprocessing to extract the necessary information.
\begin{itemize}
    \item \textbf{Deleted incorrect data rows}: While mining data, there are few incorrect rows due to the process of collecting and storing data. We decided to delete these rows from the data set.
    \item \textbf{Filled missing value}: To deal with missing values, we fill them with different strategies: numbers with 0, timestamps with the min timestamp and post messages with empty string
    \item \textbf{Extracted date time features from timestamp values}: For each timestamp value, we decoded these to date time values to enrich feature: minutes, hours, days, months, years, weekdays, etc.
    \item \textbf{Created \texttt{user\_score} feature}: For user id, we created a user reputation score metric based on previous posts in dataset. This score is used to evaluate the user's future posts
    \item \textbf{Created \texttt{image\_count} feature}: With images of each post, we compiled several information, including: number of images and image's aspect ratio
    \item \textbf{Preprocessed \texttt{post\_message} feature}: We perform post messages preprocessing more carefully than the rest. The processing stages are listed below:
    \begin{itemize}
        \item Filled missing value with empty string
        \item Standardized Vietnamese punctuation
        \item Removed HTML tags
        \item Replaced email, links, phone, numbers, emoji, date time with new corresponding token
    \end{itemize}
\end{itemize}

\subsection{Model for Tabular Data}
Metadata for the ReINTEL dataset is composed of all input features except post message (text data). We tried numerous machine learning algorithms to learn a classifier using only metadata, ranging from traditional methods: Logistic Regression, Linear Discriminant Analysis, K Nearest Neighbor, Decision Tree, Gaussian Naive Bayes, Support Vector Machine, Adaptive Boosting, Gradient Boosting, Random Forest \cite{hastie01statisticallearning}, and Extra Trees \cite{10.1007/s10994-006-6226-1} to a deep learning method: Multi-Layer Perceptron \cite{hastie01statisticallearning}

We then proceeded to select a handful of model with high performances and complexities to serve as a base model for stacking \cite{WOLPERT1992241}. Meanwhile, for the meta-model used in stacking, we chose Logistic Regression. We also did the same for blending ensemble \cite{Sill2009FeatureWeightedLS}. 

\subsection{Deep learning-based Content Classification}
\begin{figure}
    \centering
    \includegraphics[width=70mm]{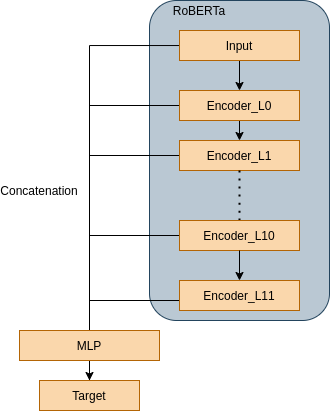}
    \caption{The architecture model for content classification using RoBERTa pre-trained model.}
    \label{fig:phobert_concat}
\end{figure}


BERT's layers capture a rich hierarchy of linguistic information, with surface features at the bottom, general syntactic knowledge in the middle, and specific semantic information at the top layer \cite{jawahar-etal-2019-bert}. Therefore, in order to better benefit for our downstream task, we incorporate as much as possible different kinds of information from our model backbone PhoBERT by concatenating [CLS] hidden states from each of 12 blocks, followed by a straightforward custom head, which is a multilayer perceptron with Dropout \cite{10.5555/2627435.2670313}. The architecture of the model is shown in the \autoref{fig:phobert_concat}.

\subsection{Deep Multi-input Model}
\begin{figure}
    \centering
    \includegraphics[width=70mm]{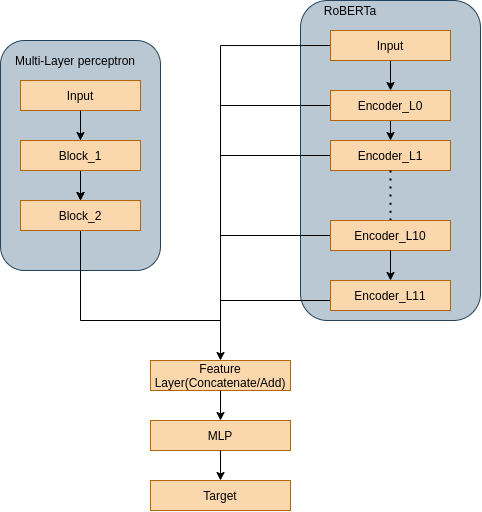}
    \caption{An illustration of our proposed deep multi-input architecture.}
    \label{fig:multimodal}
\end{figure}
Our experiments (details are in the below section) indicates that meta data is informative predictors for reliable and unreliable news classification. Therefore, we decided to combine both text and meta data to resolve the task. The structure of our multi-input model is described (in \autoref{fig:multimodal}) as followed: output features of Multi-Layer Perceptron and RoBERTa models, after being concatenated or added together, were simply passed through a custom head classifier.

\section{Experiments}
\subsection{Model Settings}
 We divide the dataset into a training set and a validation set with 10-fold cross validation method. Each fold, we use AdamW \cite{kingma2014adam} for optimization with a learning rate of $10^{-5}$ and a batch size of 32. Warm-up learning was applied, with the chosen maximum learning rate was $2\times 10^{-5}$. Except for all bias parameters and coefficients of LayerNorm layers \cite{ba2016layer}, the rest of the model's parameters were regularized with weight decay to reduce overfitting. We used a regularization coefficient of 0.01. The number of training epochs was 20.
 
 Instead of using cross-entropy loss, we implemented a label smoothing cross-entropy loss function, a combination of cross-entropy loss and label smoothing \cite{mller2019does}. The smoothing rate is set to 0.15. 
\subsection{Fine-tuning technique}
We applied state-of-the-art fine-tuning techniques including: gradual unfreezing, discriminate learning rate, warm-up learning rate schedule \cite{inproceedings} to perform effective task adaptation \cite{gururangan2020dont}. 
\subsection{Training Strategies}
We apply four training strategies to study the effects of combining text and mate data on our above-mentioned multi-data model's performance. Notice here that we used the pre-trained weights of RoBERTa as the initialization for the textual-feature-extraction-model's backbone in all strategies. We refer to the textual and meta feature extraction parts of the multi-source model are referred as text and meta submodel for short. Our training policies are described as followed:
\begin{itemize}
    \item Strategy 1 (S1): The parameters of both the text submodel's head and the meta submodel are initialized randomly
    
    \item Strategy 2 (S2): The meta submodel will be trained for the task first. Its feature extraction part (all layers except the output one used for classification) is used to combine with the text submodel. The parameters of the text submodel's head are initialized randomly.
    \item Strategy 3 (S3): Meta submodel is un-trained when incorporates with the text submodel, which is already fine-tuned with the task.
    
    \item Strategy 4 (S4): Both the two submodels are trained/fine-tuned with the classification task before being combined for further training.
\end{itemize}
\subsection{System configuration}
Our experiments are conducted on a computer with Intel Core i7 9700K Turbo 4.9GHz, 32GB of RAM, GPU GeForce GTX 2080Ti, and 1TB SSD hard disk. 
\section{Experimental Results}

\subsection{Evaluation metrics}
For this work, we used the Area Under the Receiver Operating Characteristic Curve (ROC-AUC), a common evaluation metrics for classification tasks.
The Receiver Operating Characteristic (ROC) curve shows how well a model classify samples by plotting the true positive rate against the false positive rate at various thresholds. To turn the graph into a numerical metrics, the Area Under Curve (AUC) is then evaluated. A maximum value of 1.0 indicates that the model predicts correctly for all thresholds, and a minimum of 0.0 implies the model gets everything wrong all the time. The formula for ROC-AUC is
\begin{equation}
 \text{ROC-AUC} =  \displaystyle \int_0^{+\infty} \int_{-\infty}^{+\infty} {f_1(u)f_0(u-v)}dudv
\end{equation}
 where $f_1$ and $f_0$ are the density functions. \\
\subsection{Our results}
Our results are shown in Table 
\ref{table:experimentresults3}
\ref{table:experimentresults}
\ref{table:experimentresults1}
\ref{table:experimentresults2}


\begin{table}[t]
\centering 
\caption{Performance of models using only meta data.}
\begin{tabular}{lc}
\toprule 
Method    & ROC-AUC     \\ \midrule 
Logistic Regression & 0.545037                \\
Linear Discriminant Analysis & 0.545037 \\
K Nearest Neighbors & 0.633251 \\
Decision Tree & 0.657217 \\
Gaussian Naive Bayes & 0.588978 \\
Support Vector Machine & 0.599256 \\
Adaptive Boosting & 0.673511 \\
Gradient Boosting & \textbf{0.733850} \\
Random Forest & 0.727192 \\
Extra Tree & 0.651323 \\
Multi-Layer Perceptron & 0.604653 \\\bottomrule
\end{tabular}
\label{table:experimentresults3}
\end{table}

\begin{table}[]
\centering
\caption{ROC-AUC score on public test of combining feature from blocks. Input model is the text content of the news.}
\begin{tabular}{ll}
\toprule
Blocks     & ROC-AUC    \\\midrule
Block 1-6  & 0.913251   \\
Block 6-12 & 0.937330   \\
Block 9-12 & 0.921147    \\
Block 1-12 & 0.939915  \\
Block 1-12 (Ensemble)   & \textbf{0.941811} \\\bottomrule
\end{tabular}
\label{table:experimentresults}
\end{table}




\begin{table}[]
\centering
\caption{Performance of models using only either text or meta data.}
\begin{tabular}{ll}
\toprule
Blocks     & ROC-AUC    \\\midrule
Only meta data   & 0.7338   \\
Only text data  & \textbf{0.9628}  \\ \bottomrule
\end{tabular}
\label{table:experimentresults1}
\end{table}

Table \ref{table:experimentresults3} compares the effectiveness of traditional machine learning algorithm on metadata. The performance ranges from a ROC-AUC score of 0.5450 with a simple Logistic Regression, to 0.7338 through employing Gradient Boosting across various models. Despite achieving results not as competitive as which of Gradient Boosting, the Multi-Layer Perceptron model was chosen due to its differentiability, which enabled joint training with the textual model (details in Section 3.5).
Most of the aforementioned model's performances are significantly better random guessing, indicating that metadata is an informative predictor for the news classification task.

Table \ref{table:experimentresults} shows the ROC-AUC scores as we tried incorporating different embeddings from different RoBERTa blocks. Specifically, as illustrated in \autoref{fig:phobert_concat}, we selected a subset of all embeddings RoBERTa generated, which are then concatenated together and passed through a classifier. Amongst our trials, an ensemble of various combinations across all embeddings achieved the highest AUC-ROC score of 0.9418.


Table \ref{table:experimentresults1} highlights one of the major discoveries of our work. It presents our best results for models using only meta- or text data to classify SNS. The performance gap between the two models is significant (more than 0.20 in ROC-AUC score), pointing out that textual features are more predictive than metadata. Besides, using only meta-features is considerably more accurate than random guess (0.7338 ROC-AUC score), indicating that its information can be employed to train a better model.

\begin{table}[]
\centering
\caption{Performances of multi-data model with different training strategies.}
\begin{tabular}{ll}
\toprule
Blocks     & ROC-AUC    \\\midrule
Strategy 1 (S1)  & 0.9058   \\ 
Strategy 2 (S2) & 0.9399  \\ 
Strategy 3 (S3) & 0.9552      \\ 
Strategy 4 (S4) & \textbf{0.9628}   \\ \bottomrule 
\end{tabular}
\label{table:experimentresults2}
\end{table}

 Table \ref{table:experimentresults2} sheds lights on how to effectively combined multi-source data. S1, S2, S3, and S4 in the table refer to the previously-mentioned strategy 1, strategy 2, strategy 3, and strategy 4. S1 and S2 result in the least performance among the four, less than almost 0.05 and 0.02 ROC-AUC score than our second best strategies, S4. Additionally, compared to training with only textual features even better than S1 and inconsiderably worse than S2.  This result indicates that fine-tuning text submodel with the task before combining with meta submodel is crucial to achieving high performance. 
 
The worsen results of S1 compared to S2 and S3 compared to S4 points out that pretraining meta submodel before the combination of 2 submodels enhances the overall training.



\section{Conclusion}
This paper has constructed a comprehensive comparative study to discover the effectiveness of models with multiple inputs and mixed data. We have explored and proposed different training strategies to train the hybrid deep neural architecture for reliable intelligence identification task. By conducting experiments using PhoBERT, we have demonstrated that combining mixed data with particular training plans leads to our best results. With our proposed methods, we have achieved a competitive performance of 94.18\% ROC-score on the public test and 94.62\% ROC-score on the private test set in VLSP's ReINTEL 2020 campaign.

\bibliography{vlsp2020}
\bibliographystyle{acl_natbib}

\end{document}